\documentclass[sigconf]{acmart}
\AtBeginDocument{%
  }

\usepackage{multirow}
\usepackage{booktabs}
\usepackage{array}
\usepackage{makecell}
\usepackage{colortbl}
\usepackage{siunitx}
\usepackage{float}
\usepackage{pifont}
\usepackage{bm}  
\usepackage[table]{xcolor}
\definecolor{lavender}{rgb}{0.9, 0.9, 0.98}
\usepackage{amsmath}

\usepackage{amssymb}
\usepackage{algorithm}
\usepackage{algpseudocode}
\usepackage{fontawesome} 
\definecolor{lightblue}{RGB}{230,240,255}
\definecolor{lightgray}{RGB}{245,245,245}
\definecolor{lightyellow}{RGB}{255,250,230}
\definecolor{bestcolor}{RGB}{220,255,220}
\definecolor{ForestGreen}{RGB}{34,139,34}
\definecolor{BrickRed}{RGB}{178,34,34}
\usepackage{tcolorbox}
\usepackage{colortbl}
\usepackage{color}
\usepackage{xcolor}
\usepackage{mdframed}
\definecolor{NavyBlue}{rgb}{0.0, 0.0, 0.5}
\definecolor{BlueGreen}{rgb}{0.0, 0.5, 0.5}
\definecolor{Plum}{rgb}{0.5, 0.0, 0.5}
\definecolor{ForestGreen}{rgb}{0.0, 0.26, 0.15}
\definecolor{Red}{rgb}{1.0, 0.0, 0.0}
\usepackage{tabularx}
\usepackage{tabularx}
\usepackage{hyperref}

\copyrightyear{2025}
\acmYear{2025}
\setcopyright{acmlicensed}
\acmConference[MM '25] {Proceedings of the 33rd ACM International Conference on Multimedia}{October 27--31, 2025}{Dublin, Ireland.}
\acmBooktitle{Proceedings of the 33rd ACM International Conference on Multimedia (MM '25), October 27--31, 2025, Dublin, Ireland}
\acmISBN{979-8-4007-2035-2/2025/10}
\acmDOI{10.1145/3746027.3755731}

\begin{document}

\title{Draw with Thought: Unleashing Multimodal Reasoning for Scientific Diagram Generation}

\author{Zhiqing Cui}
\authornote{Both authors contributed equally to this research.}
\email{zhiqing@nuist.edu.cn}
\affiliation{%
  \institution{Nanjing University of Information Science \& Technology}
  \city{Nanjing}
  \country{China}
}

\author{Jiahao Yuan}
\email{jhyuan.cs@gmail.com}
\authornotemark[1]
\affiliation{%
  \institution{East China Normal University}
  \city{Shanghai}
  \country{China}}

\author{Hanqing Wang}
\email{wanghanqing0424@gmail.com}
\affiliation{%
  \institution{The Hong Kong University of Science and Technology (Guangzhou)}
  \city{Guangzhou}
  \country{China}
}
\author{Yanshu Li}
\email{yanshu_li1@brown.edu}
\affiliation{%
  \institution{Brown University}
  \city{Providence}
  \country{America}
}
\author{Chenxu Du}
\email{dcx_swjtu@outlook.com}
\affiliation{%
 \institution{Southwest Jiaotong University}
 \city{Chengdu}
 \country{China}}

 \author{Zhenglong Ding}
 \authornote{Corresponding Author.}

\email{zlding@nuist.edu.cn}
\affiliation{%
  \institution{Nanjing University of Information Science \& Technology}
  \city{Nanjing}
  \country{China}
}

\renewcommand{\shortauthors}{Zhiqing Cui et al.}
\begin{abstract}
Scientific diagrams are vital tools for communicating structured knowledge across disciplines. However, they are often published as static raster images, losing symbolic semantics and limiting reuse. While Multimodal Large Language Models (MLLMs) offer a pathway to bridging vision and structure, existing methods lack semantic control and structural interpretability, especially on complex diagrams. We propose Draw with Thought (DwT), a training-free framework that guides MLLMs to reconstruct diagrams into editable mxGraph XML code through cognitively inspired Chain-of-Thought reasoning. DwT enables interpretable and controllable outputs without model fine-tuning by dividing the task into two stages: Coarse-to-Fine Planning, which handles perceptual structuring and semantic specification, and Structure-Aware Code Generation, enhanced by format-guided refinement. To support evaluation, we release Plot2XML, a benchmark of 247 real-world scientific diagrams with gold-standard XML annotations. Extensive experiments across eight MLLMs show that our approach yields high-fidelity, semantically aligned, and structurally valid reconstructions, with human evaluations confirming strong alignment in both accuracy and visual aesthetics, offering a scalable solution for converting static visuals into structurally valid and renderable representations and advancing machine understanding of scientific graphics.

\end{abstract}



\begin{teaserfigure}
    \centering
    \includegraphics[width=0.95\textwidth]{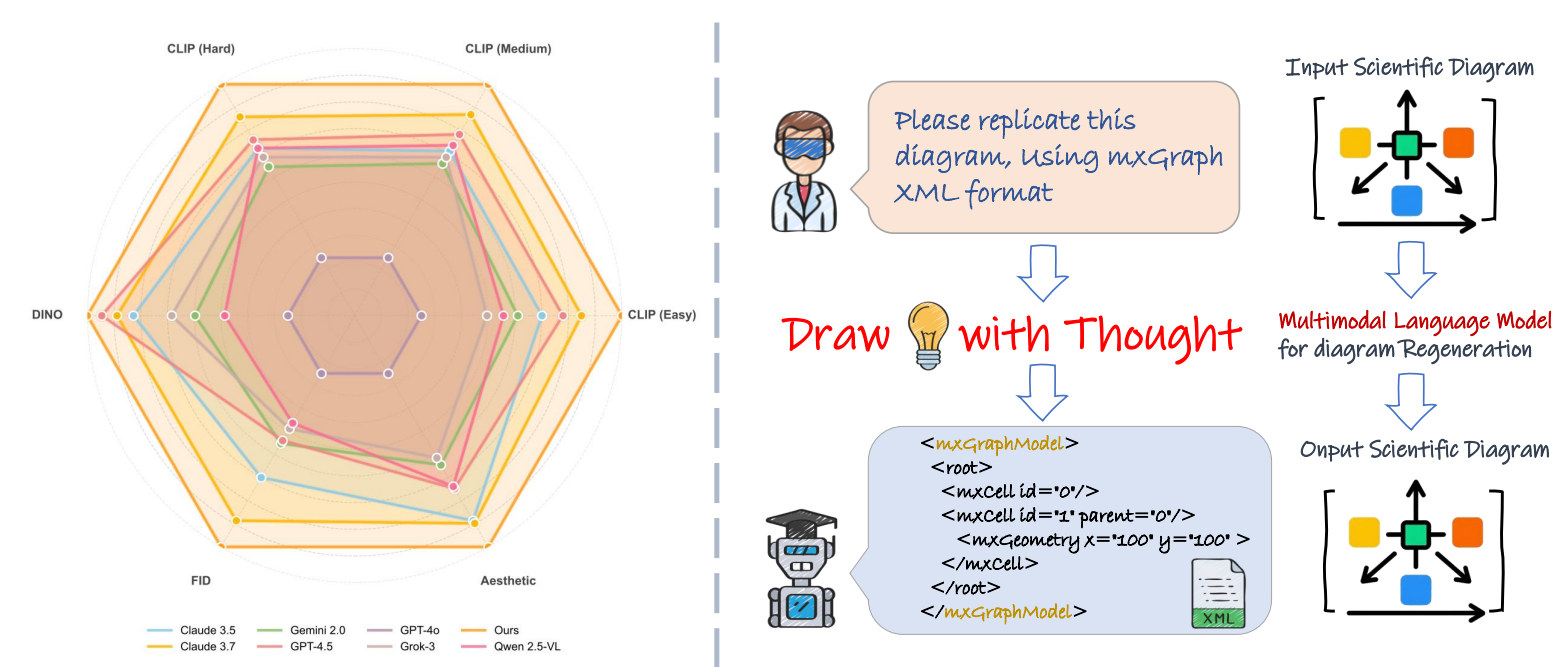}
    \caption{Overview of our Draw with Thought framework.}
    \label{fig:framework_overview}
\end{teaserfigure}

\maketitle

\section{Introduction}
Scientific diagrams, such as model architectures, system workflows, and algorithmic flowcharts, are foundational to scholarly communication, providing compact visual representations of structured ideas \cite{hegarty1991diagrams}. However, most diagrams are published as rasterized PDF or PNG images \cite{xing2024empowering}, discarding their underlying symbolic structure and making them non-editable, non-executable, and difficult to interpret or reuse programmatically \cite{chen2024onechart, favreau2016fidelity, wang2021deepvecfont}.

\begin{figure*}[h!]
  \centering
  \includegraphics[width=0.91\linewidth]{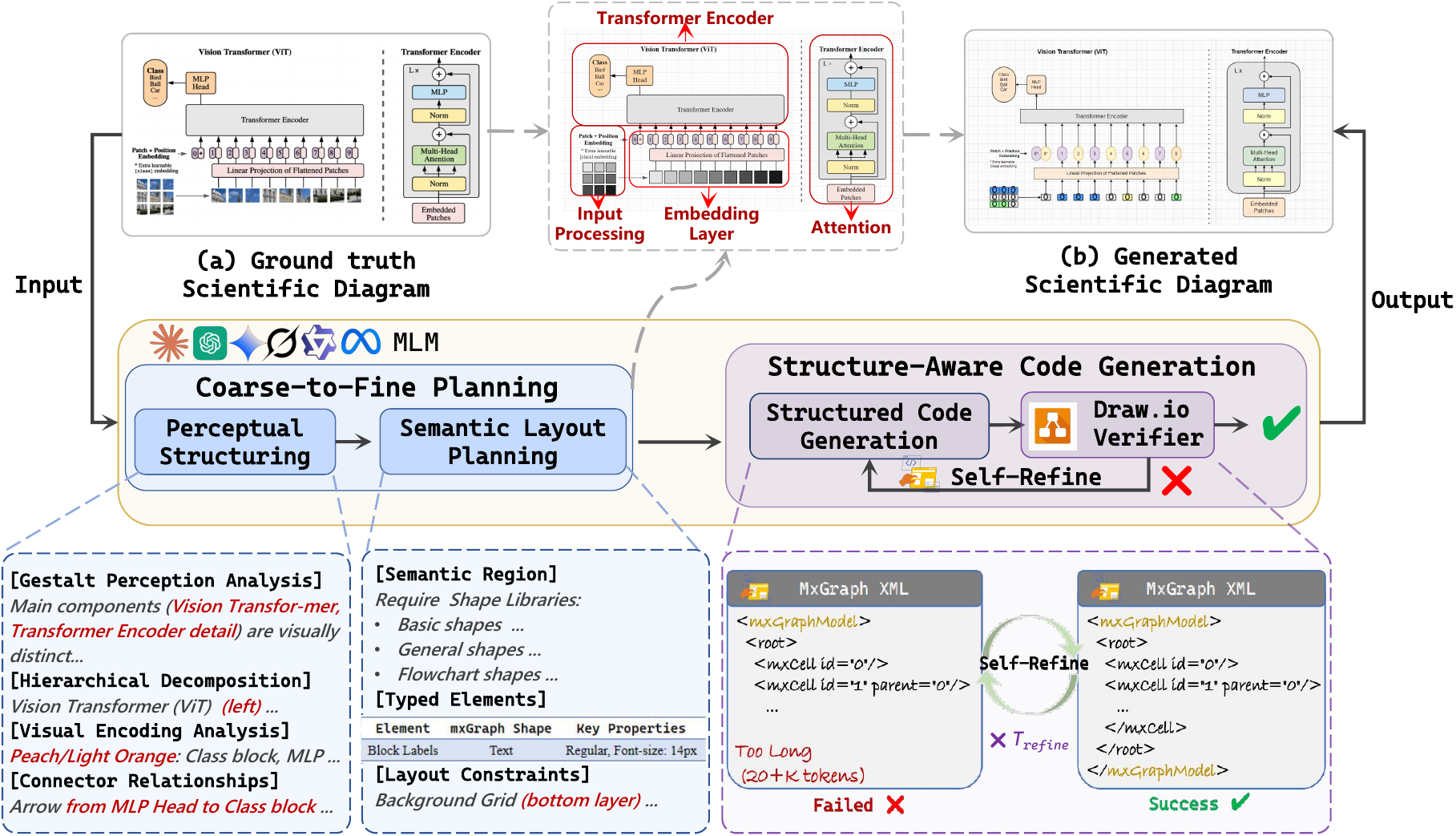}
  \caption{Overview of our Draw with Thought. The framework for scientific diagram generation processes input diagrams through Coarse-to-Fine Planning (perceptual structuring and semantic layout planning) followed by Structure-Aware Code Generation to produce editable, high-fidelity diagram representations.} 
  \label{fig:method}
\end{figure*}

Recovering structured representations from such diagrams entails mapping raw visual inputs to symbolic layouts comprising nodes, edges, spatial relations, and logical groupings \cite{belouadi2024detikzify}—a task that lies at the intersection of visual perception and structured abstraction. While Vision-Language Models (VLMs) have shown strong capabilities in aligning images with structured outputs, as demonstrated in captioning \cite{li2023blip, NEURIPS2023_9a6a435e}, visual QA \cite{song2022clip}, and diagram layout prediction \cite{roberts2024image2struct}, their application to scientific diagram parsing remains nascent and largely limited to visually regular or low-complexity inputs. Existing studies have adopted various target formats that differ in expressiveness and abstraction level, such as SVG for vector rendering \cite{rodriguez2023starvector,xing2024empowering}, TikZ for LaTeX-integrated graphics \cite{belouadiautomatikz,belouadi2024detikzify}, and Python-based code for chart generation \cite{wu2024plot2code}. While SVG enables resolution-independent rendering, its reliance on low-level geometric primitives limits semantic interpretability \cite{belouadi2024detikzify}; TikZ supports precise layout control but introduces parsing challenges due to procedural syntax and structural variability \cite{belouadiautomatikz,belouadi2024detikzify}; and although Python formats are effective for domain-specific plots, they lack the modularity and compositional flexibility needed to capture the heterogeneity of general-purpose scientific diagrams \cite{wu2024plot2code}. In contrast, XML-based formats such as mxGraph offer an explicit and editable graph abstraction, encoding diagrams as structured layouts with semantically grounded nodes, edges, and geometric constraints. This makes them particularly well-suited as target representations for recovering the symbolic structure of scientific diagrams from raw visual inputs \cite{roberts2024image2struct}.

In summary, existing methods for scientific diagram parsing face three key limitations: (1) reliance on shallow or domain-specific formats that restrict structural expressiveness \cite{xing2024empowering,rodriguez2023starvector}; (2) modeling paradigms that focus on layout detection or retrieval, lacking symbolic reasoning and structural supervision \cite{xing2024empowering}; and (3) assumptions of clean vector inputs or templates \cite{wu2024plot2code}, which hinder generalization to real-world, heterogeneous, rasterized diagrams.

To address these limitations, we present \textit{Draw with Thought}, a training-free framework that leverages Multimodal Large Language Models (MLLMs) for symbolic reconstruction of scientific diagrams. Our approach draws inspiration from cognitive load theory \cite{plass2010cognitive}, which suggests that breaking down complex tasks into manageable steps facilitates reasoning, and from structure mapping theory \cite{gentner1983structure}, which models analogy-making as a process of aligning relational structures. Building on such insights, we design a cognitively inspired Chain-of-Thought (CoT) prompting strategy that guides MLLMs through two key stages: \textbf{Coarse-to-Fine Planning} (Section~\ref{sec:perceptual}), which includes Perceptual Structuring and Semantic Specification, and \textbf{Structure-Aware Code Generation} (Section~\ref{sec:symbolic}), which incorporates iterative Format-Guided Refinement. These stages reflect how humans process diagrams—by sequentially recognizing visual elements, inferring their relationships, and assembling them into structured representations. Unlike prior works that rely on vision-only perception  \cite{wu2024plot2code} or heuristic-based templates \cite{fang2025got}, our framework introduces a unified, interpretable reasoning pipeline compatible with diverse diagram styles and semantic structures. It supports high-fidelity generation of editable mxGraph XML code without any model fine-tuning. Our key contributions include:
\begin{itemize}
  \item We introduce \textbf{Draw with Thought}, which guides MLLMs to generate executable mxGraph representations from rasterized scientific diagrams via cognitively inspired chain-of-thought prompting: covering visual decomposition, relational inference, and symbolic code synthesis.
  \item We construct \textbf{Plot2XML}, a benchmark consisting of 247 complex diagrams. It spans key domains such as computer vision, multimodal model, and broader AI systems, covering most of the key use cases for scientific diagrams.
  \item Comprehensive experiments across multiple MLLMs demonstrate that our approach achieves high fidelity, semantic alignment, and easy editability—all without the need for model fine-tuning.
\end{itemize}

\section{Related Work}
\subsection{Image-to-Markup Conversion} 
Image-to-markup conversion supports a wide range of diagram types. Existing Studies rely on low-level vector primitives like SVG \cite{rodriguez2023starvector, xing2024empowering}, which ensure resolution independence but offer limited semantic detail. Others use Python-based chart-generation scripts including Plot2Code \cite{wu2024plot2code}, PandasPlotBench \cite{galimzyanov2024drawing} and Chart2Code \cite{zhao2025chartcoder}, which excel in producing domain-specific plots yet struggle with heterogeneous diagrams \cite{belouadi2024detikzify}. Math-centric methods translate geometric figures or handwritten equations into \LaTeX\ code \cite{blechernougat, wang2020pdf2latex} and address symbolic inference, but often remain confined to specialized content. Besides, existing datasets largely fall short in covering key dimensions required for structured scientific diagram understanding. SciGraphQA \cite{li2023scigraphqa} lacks visual inputs and complex structure, Design2Code \cite{si2024design2code} focuses on UI layouts with HTML rendering but lacks diagram-level abstraction, SketchFig \cite{belouadi2024detikzify} includes hand-drawn diagrams but provides limited layout semantics and MMCode \cite{li2024mmcode} supports multimodal prompts but lacks authentic scientific content and XML-level structure. In contrast, XML-based structured markup language \cite{roberts2024image2struct} under mxGraph standards \cite{kaplan2024bridging}, captures nodes, edges, and layout constraints in a unified form—an especially relevant feature for complex scientific diagrams that involve diverse graphical elements and rich relational structures. Consequently, to handle the intricacies of scientific diagram generation \cite{hegarty1991diagrams}, our work adopts mxGraph to define the scientific diagram parsing task as an image-to-XML pipeline, enabling more expressive and editable representations. 

\subsection{Multimodal Large Language Models in Code Generation}
Recent advances in Multimodal Large Language Models (MLLMs), such as Qwen-vl \cite{bai2025qwen2} and GPT-4 \cite{hurst2024gpt} demonstrate strong abilities to extract structured meaning from complex visual inputs, achieving notable results in image captioning \cite{NEURIPS2023_9a6a435e, li2023blip}, visual question answering \cite{song2022clip}, and scene-level reasoning \cite{gao2024cantor}. Meanwhile, code-oriented LLMs, including InCoder \cite{friedincoder}, StarCoder \cite{listarcoder}, and Code Llama \cite{grattafiori2023code}, excel in generating executable programs across multiple languages\cite{zhang2024scimage,yang2024chartmimic}. Existing studies that integrate visual cues with textual embeddings \cite{xing2024empowering, roberts2024image2struct} often focus on simpler diagrams or charts \cite{wu2024plot2code}, where step-by-step reasoning involving Chain-of-Thought \cite{wei2022chain}, self-refine \cite{madaan2023self, fang2025got}, self-reflect \cite{cheng2024vision} reasoning-and-acting \cite{yaoreact,wu2025webwalker}and plan-and-solve \cite{wang2023plan,qiao2024agent} can enhance interpretability. However, many approaches rely on standardized domains \cite{xing2024empowering} and struggle to capture the rich semantics of real-world scientific diagrams \cite{belouadi2024detikzify}. Finetuning models to generate lengthy code like XML \cite{xie2024wukong}, SVG \cite{rodriguez2023starvector,xing2024empowering} or LaTeX \cite{belouadi2024detikzify} introduces substantial computational overhead, especially for small-parameter LLMs that must handle highly varied and domain-specific content. To overcome these challenges and facilitate XML-based code generation from complex scientific images, our work adopts a training-free strategy that leverages MLLMs for robust spatial reasoning, domain-specific logic, and knowledge-intensive tasks.

Inspired by cognitive load theory \cite{plass2010cognitive}, which advocates breaking intricate problems into manageable steps-and structure mapping theory \cite{gentner1983structure}—which models human reasoning through the alignment of relational structures—we propose \textit{Draw with Thought}, a framework that guides MLLMs to parse and synthesize diagrammatic information without the need for expensive model finetuning.

\begin{table*}[t]
\centering
\renewcommand{\arraystretch}{1.2}
\setlength{\tabcolsep}{5pt}
\caption{Comparison of Scientific Diagram Datasets for Structural Understanding and Representation. We evaluate datasets across multiple dimensions. Full support (\textcolor{green!70!black}{\Large\bm{$\checkmark$}}) indicates comprehensive capability, partial support (\textcolor{orange!80!black}{\Large$\checkmark$}) indicates limited capability, and no support (\textcolor{red!70!black}{\Large\bm{$\times$}}) indicates absence of capability. Multimodality notation: T: Text-only, I: Image-only, I+T: Multi-modal with both images and text. Data authenticity reflects whether datasets contain real-world scientific diagrams from research publications and the complexity metric is expressed as the difficulty of data generation and evaluation.}
\resizebox{0.95\linewidth}{!}{
\begin{tabular}{l|cc|ccc|c|c}
\toprule
\multirow{2}{*}{\textbf{Dataset}} & \multicolumn{2}{c|}{\textbf{Domain Characteristics}} & \multicolumn{3}{c|}{\textbf{Multimodality}} & \multirow{2}{*}{\textbf{Format}} & \multirow{2}{*}{\textbf{Complexity}} \\
\cmidrule(lr){2-3} \cmidrule(lr){4-6}
 & Figure Type & Data Authenticity & T & I & I+T & & \\
\midrule
SciGraphQA\cite{li2023scigraphqa} & Scientific Charts & \textcolor{red!70!black}{\Large\bm{$\times$}} & \textcolor{green!70!black}{\Large\bm{$\checkmark$}} & \textcolor{red!70!black}{\Large\bm{$\times$}} & \textcolor{red!70!black}{\Large\bm{$\times$}} & Text & \textcolor{red!70!black}{\Large\bm{$\times$}} \\
Design2Code\cite{si2024design2code} & Website UI & \textcolor{green!70!black}{\Large\bm{$\checkmark$}} & \textcolor{red!70!black}{\Large\bm{$\times$}} & \textcolor{green!70!black}{\Large\bm{$\checkmark$}} & \textcolor{red!70!black}{\Large\bm{$\times$}} & HTML & \textcolor{orange!80!black}{\Large$\checkmark$} \\
SketchFig\cite{belouadi2024detikzify} & Hand-drawn Diagrams & \textcolor{orange!80!black}{\Large$\checkmark$} & \textcolor{red!70!black}{\Large\bm{$\times$}} & \textcolor{green!70!black}{\Large\bm{$\checkmark$}} & \textcolor{red!70!black}{\Large\bm{$\times$}} & TikZ & \textcolor{red!70!black}{\Large\bm{$\times$}} \\
MMCode\cite{li2024mmcode} & Algorithm Diagrams & \textcolor{red!70!black}{\Large\bm{$\times$}} & \textcolor{green!70!black}{\Large\bm{$\checkmark$}} & \textcolor{green!70!black}{\Large\bm{$\checkmark$}} & \textcolor{orange!80!black}{\Large$\checkmark$} & Code & \textcolor{orange!80!black}{\Large$\checkmark$} \\
\midrule
\textbf{Plot2XML} & \textbf{Complex Research Diagrams} & \textcolor{green!70!black}{\Large\bm{$\checkmark$}} & \textcolor{green!70!black}{\Large\bm{$\checkmark$}} & \textcolor{green!70!black}{\Large\bm{$\checkmark$}} & \textcolor{green!70!black}{\Large\bm{$\checkmark$}} & \textbf{XML} & \textcolor{green!70!black}{\Large\bm{$\checkmark$}} \\
\bottomrule
\end{tabular}
}
\label{tab:dataset_comparison}
\end{table*}

\section{Plot2XML: A Benchmark for Scientific Diagram-to-Code Generation}
As shown in Table~\ref{tab:dataset_comparison}, prior datasets fall short in authenticity or structure, while Plot2XML provides real scientific diagrams with verified XML annotations.

\subsection{Data Collection and Curation}

We introduce Plot2XML, a novel benchmark dataset consisting of 247 scientific diagrams carefully selected from influential conference papers across multiple domains. The collection spans model architecture diagrams, flowcharts, system pipelines, and conceptual frameworks essential for scientific communication. Unlike SVG representations\cite{eisenberg2014svg,song2025layertracer}, our Plot2XML diagrams in mxGraph format exhibit a geometric increase in both element count and code complexity, making this the first large-scale collection of scientific diagrams with manually verified markup-based structural annotations. 


\subsection{Complexity Analysis}
\begin{figure}[htbp]
    \centering
    \includegraphics[width=0.92\columnwidth]{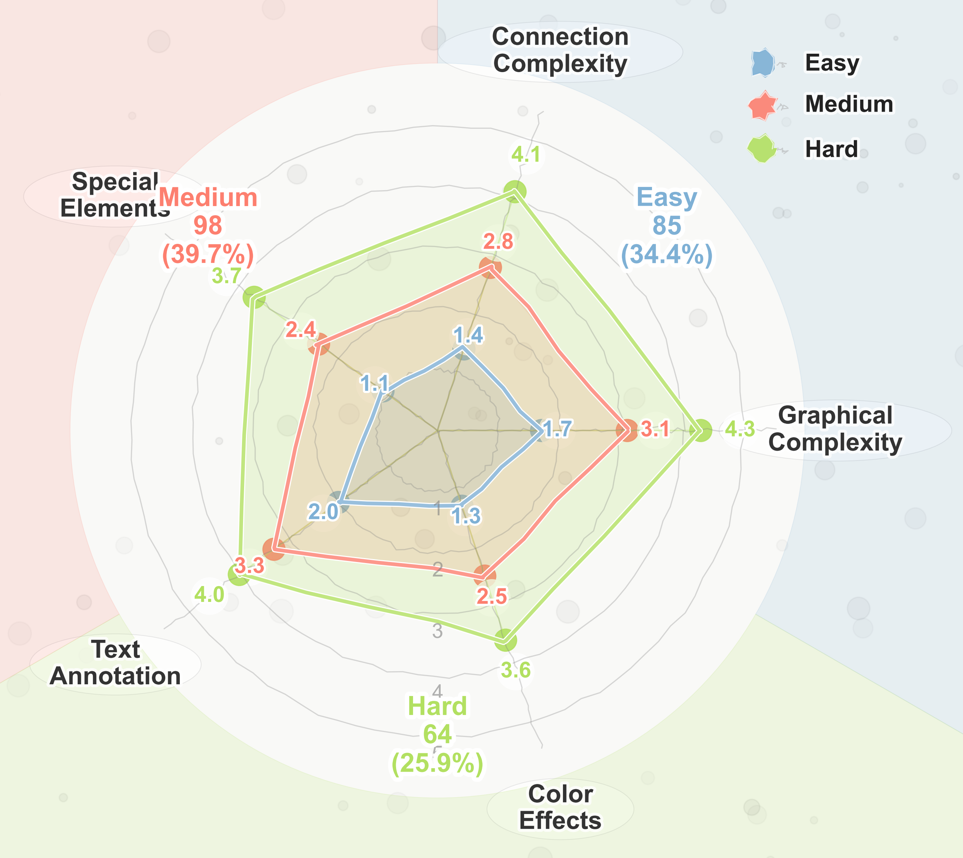}
    \caption{Complexity analysis of scientific diagrams. Visualized as a radar chart over five dimensions: Connection Complexity, Graphical Complexity, Color Effects, Text Annotation, and Special Elements. The chart compares three difficulty levels—\textcolor[HTML]{7eb0d5}{Easy} (blue, 85 diagrams, 34.4\%), \textcolor[HTML]{fd7f6f}{Medium} (coral, 98 diagrams, 39.7\%), and \textcolor[HTML]{b2e061}{Hard} (green, 64 diagrams, 25.9\%)—highlighting distinct structural patterns across complexity axes.}
    \label{fig:diagram_complexity}
\end{figure}

To systematically categorize the diagrams, we developed a multi-dimensional complexity analysis framework evaluating each diagram across five key dimensions: Connection Complexity (relationships between elements), Graphical Complexity (visual sophistication), Color Effects (semantic encoding), Text Annotation (textual density), and Special Elements (non-standard visual components). Based on these dimensions, we classified diagrams into three difficulty levels. As shown in Figure~\ref{fig:diagram_complexity}, Hard diagrams exhibit consistently higher complexity scores across all dimensions, with particularly pronounced differences in Connection Complexity ($score = 4.1$) and Graphical Complexity ($score = 4.3$). 

\begin{algorithm}[t]
\caption{\textsc{Draw with Thought (DwT)}}
\label{alg:dwt}
\begin{algorithmic}[1]
\Require Rasterized scientific diagram $D_{scientific}$
\Ensure Structured code $Y_{mxGraph}$ in \texttt{mxGraph XML}

\State \textbf{Stage I: Coarse-to-Fine Planning for
Structured Visual Understanding} \hfill \Comment{(Sec.~\ref{sec:perceptual})}

\State \quad \textit{Phase 1.1: Perceptual Structuring}
\State \quad $T_{percept} = \mathcal{M}_{MLLM}(D_{scientific},{Thought}_{percept})$
\State \quad \quad $= (T_{gestalt}, T_{hierarchy}, T_{encoding}, T_{connector})$
\State {\footnotesize\textcolor{darkgray}{\quad $\triangleright$ $T_{gestalt}$: perceptual grouping via Gestalt principles}}
\State {\footnotesize\textcolor{darkgray}{\quad $\triangleright$ $T_{hierarchy}$: decomposition into visual primitives}}
\State {\footnotesize\textcolor{darkgray}{\quad $\triangleright$ $T_{encoding}$: visual-to-semantic variable mapping}}
\State {\footnotesize\textcolor{darkgray}{\quad $\triangleright$ $T_{connector}$: element linkage and routing topology}}

\State \quad \textit{Phase 1.2: SemanticSpecification}
\State \quad $T_{hierarchy} = \mathcal{M}_{MLLM}(T_{percept}, {Thought}_{hierarchy})$
\State \quad \quad $= (\mathcal{R}, \mathcal{E}, \mathcal{L})$
\State {\footnotesize\textcolor{darkgray}{\quad $\triangleright$ $\mathcal{R}$: semantic regions (e.g., \texttt{Input}, \texttt{Output})}}
\State {\footnotesize\textcolor{darkgray}{\quad $\triangleright$ $\mathcal{E}$: typed elements with class labels (e.g., \texttt{Entity})}}
\State {\footnotesize\textcolor{darkgray}{\quad $\triangleright$ $\mathcal{L}$: layout constraints (alignment, layering, connectivity)}}

\State \textbf{Stage II: Structure-Aware Code Generation via
Progressive Realization} \hfill \Comment{(Sec.~\ref{sec:symbolic})}
\State \quad \textit{Phase 2.1: Initial Structured Code Generation}
\State \quad $Y_{mxGraph} = \mathcal{M}_{MLLM}(T_{hierarchy}, {Thought}_{\text{code}})$
\State \quad \quad $= \bigcup_i Y_i$ =$(Y_{doc}, Y_{style}, Y_{node}, Y_{layout}, Y_{\text{edge}})$
\State {\footnotesize\textcolor{darkgray}{\quad $\triangleright$ $Y_{\text{doc}}$: root declarations (\texttt{mxfile}, \texttt{diagram})}}
\State {\footnotesize\textcolor{darkgray}{\quad $\triangleright$ $Y_{\text{style}}$: visual style tokens (colors, borders, fonts)}}
\State {\footnotesize\textcolor{darkgray}{\quad $\triangleright$ $Y_{\text{node}}$: nodes from $\mathcal{E}$ with geometry and IDs}}
\State {\footnotesize\textcolor{darkgray}{\quad $\triangleright$ $Y_{\text{layout}}$: layout constraints from $\mathcal{L}$}}
\State {\footnotesize\textcolor{darkgray}{\quad $\triangleright$ $Y_{\text{edge}}$: directed connectors with routing logic}}
\State \quad \textit{Phase 2.2: Multi-Round Format-Guided XML Refinement}
\For{$t = 1$ \textbf{to} $T_{\text{refine}}$}
    \State $Y_{mxGraph}^{(t)} = \mathcal{M}_{MLLM}(Y_{mxGraph}^{(t{-}1)} , $ \\ \quad \quad \quad \quad \quad  \quad \quad \quad \quad \quad \quad \quad ${RefineThought}^{(t)}_{format})$
    \State \textbf{if} \texttt{Draw.ioVerifier}($Y_{mxGraph}^{(t)}$) == valid \textbf{then break}
\EndFor
\State {\footnotesize\textcolor{darkgray}{$\triangleright$ Verification ensures XML schema validity, rendering success, and structural consistency}}
\State \Return $Y_{mxGraph}^{(T^*)}$ 
\State {\footnotesize\textcolor{darkgray}{$\triangleright$ $T^* \leq T_{\text{refine}}$, earliest valid XML accepted by \texttt{Draw.io}}}
\end{algorithmic}
\end{algorithm}

\section{Draw with Thought}

Following \cite{wang2023plan}, We propose \textit{Draw with Thought} (DwT), a planning-driven framework that transforms a rasterized scientific diagram $D_{scientific}$ into structured, editable graph code $Y_{mxGraph}$ in \texttt{mxGraph} XML format by guiding a Multimodal Large Language Model $\mathcal{M}_{MLLM}$ through planning-and-generate Chain of Thought.
Inspired by how humans decompose diagrams through sequential reasoning \cite{plass2010cognitive,gentner1983structure}---from perceptual recognition to semantic understanding and symbolic execution---we explicitly separate the diagram-to-code task into two interpretable stages: \textbf{(1) Coarse-to-Fine Planning} involving ${Thought}_{percept}$ and ${Thought}_{hierarchy}$ (Section~\ref{sec:perceptual}). \textbf{(2) Structure-Aware Generation} involving ${Thought}_{code}$ and format (Section~\ref{sec:symbolic}), as detailed in Algorithm~\ref{alg:dwt} and Figure~\ref{fig:method}. This separation enables the model to reason over symbolic abstractions before emitting graphical code.

\subsection{Coarse-to-Fine Planning for Structured Visual Understanding}
\label{sec:perceptual}
\paragraph{Perceptual Structuring through Symbolic Abstraction.}To reduce perceptual complexity, we first induce a symbolic abstraction $T_{percept}$ that encodes diagram structure through layout grouping, hierarchy, visual encoding, and topology extraction. Inspired by GoT \cite{fang2025got}, we prompt Multimodal Large language Model $\mathcal{M}_{MLLM}$ to simulate perceptual reasoning via ${Thought}_{perceptual}$:
\begin{equation}
T_{percept} \sim \mathcal{M}_{MLLM}(\cdot \mid D_{scientific}, {Thought}_{percept}),
\end{equation}
We decompose the output into four complementary components:
\begin{equation}
T_{percept} = (T_{gestalt}, T_{hierarchy}, T_{encoding}, T_{connector}).
\end{equation}
where $T_{gestalt}$ encodes perceptual grouping using Gestalt principles (proximity, similarity, continuity, closure), 
$T_{hierarchy}$ defines recursive decomposition of visual objects into primitives and their nesting, $T_{encoding}$ identifies mappings from visual variables to semantic roles (e.g., color $\mapsto$ category, size $\mapsto$ quantity), and $T_{connector}$ captures all connective topology between elements, including directionality and routing.

\paragraph{Semantic Specification through Layout-Aware Abstraction}
Once perceptual reasoning is complete, the MLLM transitions to semantic abstraction, converting low-level visual cues into structured symbolic representations. Given $T_{percept}$, the model is guided by ${Thought}_{hierarchy}$ to define a semantically meaningful layout plan:
\begin{equation}
T_{hierarchy} \sim \mathcal{M}_{MLLM}(\cdot \mid T_{percept}, {Thought}_{hierarchy}),
\end{equation}

We represent the layout plan as:
\begin{equation}
T_{hierarchy} = (\mathcal{R}, \mathcal{E}, \mathcal{L}),
\end{equation}
where $\mathcal{R}$ denotes semantic regions (e.g., \texttt{Input}, \texttt{Output}), $\mathcal{E}$ is the catalog of typed visual elements with class labels and graphical roles (e.g., \texttt{Process}, \texttt{Decision}, \texttt{Entity}), and $\mathcal{L}$ formalizes spatial constraints among elements, defined as:
\begin{equation}
\mathcal{L} = \{ \texttt{align}(e_i, e_j),\ \texttt{connect}(e_i \rightarrow e_j),\ \texttt{layer}(e_k, z) \}.
\end{equation}

This abstraction step lifts diagram content from geometric perception to symbolic understanding. Importantly, the layout constraints in $\mathcal{L}$ preserve both the visual integrity and semantic logic of the diagram, serving as the backbone for code synthesis in the next stage.

\subsection{Structure-Aware Code Generation via Progressive Realization}
\label{sec:symbolic}
\paragraph{Initial Structured Code Generation.} Based on $T_{hierarchy}$, we prompt $\mathcal{M}_{MLLM}$ to progressively synthesize the final XML code $Y_{mxGraph}$:
\begin{equation}
Y_{mxGraph} \sim \mathcal{M}_{MLLM}(\cdot \mid T_{hierarchy}, {Thought}_{code}),
\end{equation}

We decompose the XML structure into five submodules:
\begin{equation}
Y^{(0)}_{mxGraph} = (Y_{doc}, Y_{style}, Y_{node}, Y_{layout}, Y_{edge}),
\end{equation}
where $Y_{doc}$ defines root-level declarations and metadata containers (e.g., \texttt{mxfile}, \texttt{diagram}, \texttt{mxGraphModel}), $Y_{style}$ encodes style dictionaries with reusable keys, $Y_{node}$ instantiates shape primitives from $\mathcal{E}$ with geometry and identifiers, $Y_{layout}$ applies $\mathcal{L}$ constraints to compute coordinates and alignment, and $Y_{edge}$ constructs directional connectors with routing points, anchors, and labels.

Finally, we define the complete output as:
\begin{equation}
\begin{aligned}
Y^{(0)}_{mxGraph} &= \bigcup_{i} Y_i, \\
&\quad \text{for } Y_i \in \{Y_{doc}, Y_{style}, Y_{node}, Y_{layout}, Y_{edge}\}.
\end{aligned}
\end{equation}

\begin{figure*}[ht!]
  \centering
  \includegraphics[width=\linewidth]{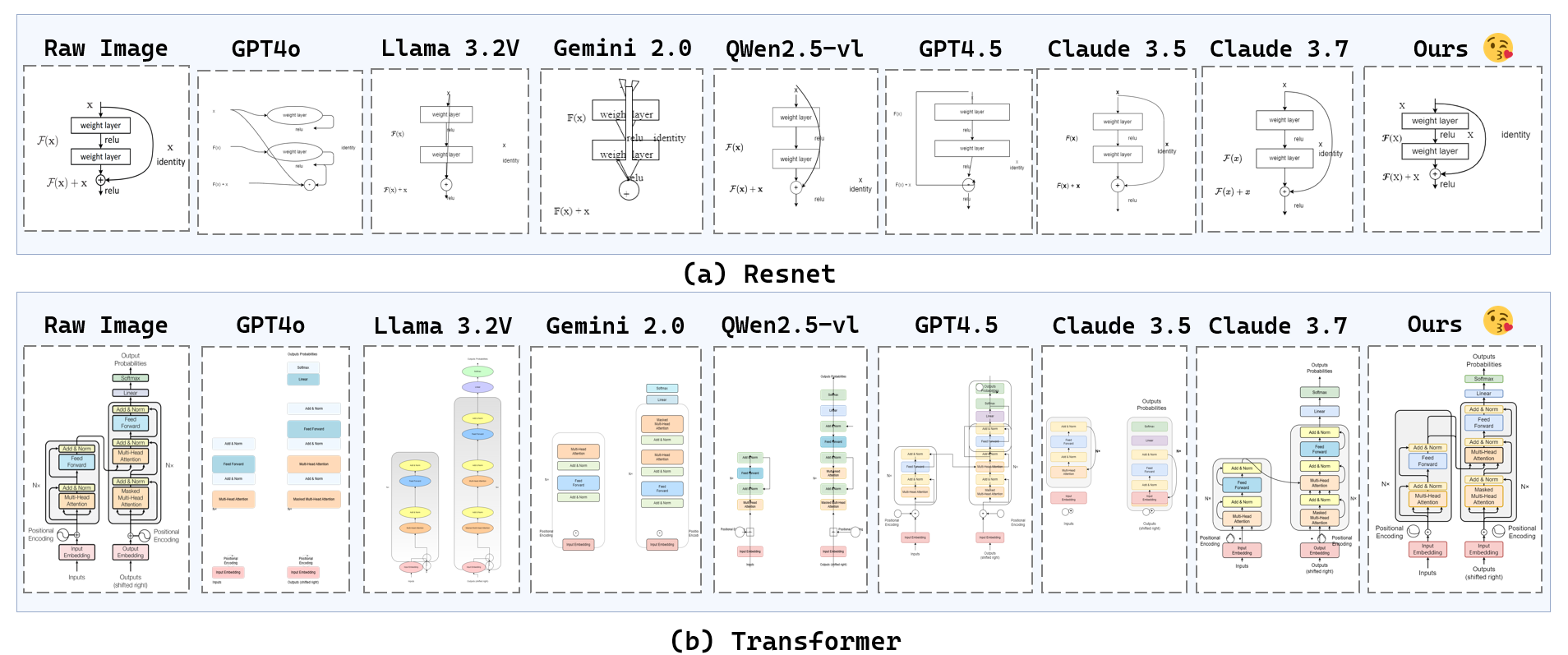}
  \caption{Qualitative Comparison Between DwT and State-of-the-Art MLLMs. We visualize the output results. We observe that our model can understand the layout and the content of the image well, achieving better performance.}
  \label{fig:case}
\end{figure*}

Each component serves a distinct role: $Y_{doc}$ initializes the document tree (e.g., \texttt{mxfile}, \texttt{diagram}); $Y_{style}$ defines reusable style tokens (e.g., color, font, stroke); $Y_{node}$ instantiates entities $\mathcal{E}$ with geometry and ID; $Y_{layout}$ applies spatial constraints from $\mathcal{L}$; and $Y_{edge}$ encodes directed connectors with routing semantics.

\paragraph{Multi-Round Format-Guided XML Refinement.}While structurally grounded, initial outputs often contain XML formatting flaws—such as unclosed tags, incorrect nesting, or inconsistent styles for long-context for output during inference \cite{bai2024longwriter}. To ensure syntactic correctness and structural executability, we introduce a multi-round refinement module guided by format-level reasoning.

At each refinement step $t$, $\mathcal{M}_{MLLM}$ revisit the prior output $Y^{(t-1)}_{mxGraph}$ and produce an updated version using format-specific reasoning via ${RefineThought}^{(t)}_{format}$ prompting:
\begin{align}
Y^{(t)}_{mxGraph} \sim\ 
& \mathcal{M}_{MLLM}\big(Y^{(t-1)}_{mxGraph},\ {RefineThought}^{(t)}_{format}\big), \notag \\
& \quad t = 1, \dots, T_{refine}
\end{align}

Each refinement round injects XML schema constraints, tag correction heuristics, and layout logic. The refinement process is terminated early if the intermediate output passes verification in \texttt{Draw.io}:
\begin{align}
\text{if } \texttt{Draw.ioVerifier}(Y^{(t)}_{\text{mxGraph}}) = \texttt{valid} \text{ then break}.
\end{align}

This ensures that the final output $Y^{(T^*)}_{\text{mxGraph}}$ is well-structured and executable. The resulting XML can be directly used in platforms such as \texttt{draw.io}, enabling downstream applications and interactive editing.

\begin{table*}[ht!]
\centering
\small
\setlength{\tabcolsep}{2.8pt}
\renewcommand{\arraystretch}{1.2}
\caption{Comprehensive performance comparison of AI models across three difficulty levels. Our visual design employs a gradient blue background (increasing intensity with difficulty) to enhance readability, with purple section headers and a teal highlight for the best-performing model. Star icons (\faStarO, \faStar) 
 indicate difficulty levels, while specialized icons represent model categories: \faDatabase  for base models, \faChain for Chain-of-Thought models, and \faMagic  for our proposed method. Performance metrics include CLIP and DINO scores measuring semantic consistency, FID score for image quality, aesthetic rating, and token consumption in thousands (K).}
\begin{tabular}{l|>{\columncolor{blue!3}}ccccc|>{\columncolor{blue!6}}ccccc|>{\columncolor{blue!9}}ccccc}
\toprule
\multirow{2}{*}[-0.2em]{\textbf{Model}} & \multicolumn{5}{c|}{\textbf{Easy Difficulty} \small\faStarO} & \multicolumn{5}{c|}{\textbf{Medium Difficulty} \small\faStar\faStarO} & \multicolumn{5}{c}{\textbf{Hard Difficulty} \small\faStar\faStar\faStar} \\
\cmidrule(lr){2-6} \cmidrule(lr){7-11} \cmidrule(lr){12-16}
 & \makecell{CLIP\\$\uparrow$} & \makecell{DINO\\$\uparrow$} & \makecell{FID\\$\downarrow$} & \makecell{Aesth.\\$\uparrow$} & \makecell{Tok.\\(K)} & 
\makecell{CLIP\\$\uparrow$} & \makecell{DINO\\$\uparrow$} & \makecell{FID\\$\downarrow$} & \makecell{Aesth.\\$\uparrow$} & \makecell{Tok.\\(K)} & 
\makecell{CLIP\\$\uparrow$} & \makecell{DINO\\$\uparrow$} & \makecell{FID\\$\downarrow$} & \makecell{Aesth.\\$\uparrow$} & \makecell{Tok.\\(K)} \\
\midrule
\rowcolor{purple!8}
\multicolumn{16}{c}{\textbf{Base Models} \small\faDatabase} \\
\midrule
\rowcolor{blue!2}
GPT-4o & 0.72 & 0.87 & 172 & 3.95 & 1.5 & 0.51 & 0.69 & 208 & 3.81 & 2.0 & 0.38 & 0.58 & 243 & 3.66 & 2.9 \\
Llama-3.2V-11B & 0.73 & 0.88 & 161 & 4.17 & 2.2 & 0.54 & 0.75 & 199 & 4.02 & 3.1 & 0.44 & 0.64 & 213 & 3.71 & 3.6 \\
\rowcolor{blue!2}
Gemini 2.0 & 0.76 & 0.89 & 146 & 4.32 & 3.0 & 0.59 & 0.81 & 181 & 4.06 & 6.0 & 0.48 & 0.73 & 223 & 3.85 & 10.5 \\
Grok 3 & 0.74 & 0.90 & 155 & 4.27 & 3.3 & 0.60 & 0.78 & 178 & 4.17 & 5.8 & 0.50 & 0.70 & 211 & 4.05 & 9.6 \\
\rowcolor{blue!2}
GPT-4.5 & 0.80 & 0.94 & 148 & 4.51 & 2.0 & 0.64 & 0.79 & 165 & 4.21 & 4.5 & 0.54 & 0.72 & 185 & 3.95 & 7.6 \\
Qwen 2.5 VL-32B & 0.75 & 0.88 & 158 & 4.49 & 1.9 & 0.62 & 0.81 & 168 & 4.34 & 3.6 & 0.52 & 0.73 & 195 & 4.17 & 6.6 \\
\rowcolor{blue!2}
Claude 3.5-sonnet & 0.78 & 0.92 & 119 & 4.82 & 1.8 & 0.61 & 0.84 & 148 & 4.41 & 4.1 & 0.52 & 0.77 & 180 & 4.15 & 8.8 \\
Claude 3.7-sonnet & \underline{0.82} & \underline{0.93} & \underline{73} & \underline{4.85} & 2.0 & \underline{0.68} & \underline{0.86} & \underline{112} & \underline{4.50} & 8.9 & \underline{0.60} & \underline{0.78} & \underline{150} & \underline{4.25} & 14.0 \\
\midrule
\rowcolor{purple!12}
\multicolumn{16}{c}{\textbf{Models with Chain-of-Thought (CoT)} \small\faChain} \\
\midrule
Qwen 2.5 VL-32B+CoT & 0.78 & 0.88 & 133 & 4.71 & 2.0 & 0.64 & 0.82 & 163 & 4.39 & 3.9 & 0.55 & 0.75 & 178 & 4.15 & 7.4 \\
\rowcolor{blue!2}
Claude 3.7-sonnet+CoT & 0.83 & 0.94 & 55 & 4.88 & 3.3 & 0.71 & 0.88 & 98 & 4.52 & 9.9 & 0.63 & 0.81 & 139 & 4.28 & 16.9 \\
\midrule
\rowcolor{purple!16}
\multicolumn{16}{c}{\textbf{Models with Our Method} \small\faMagic} \\
\midrule
Qwen 2.5 VL-32B+DwT & 0.81 & 0.93 & 67 & 4.80 & 2.1 & 0.70 & 0.84 & 129 & 4.56 & 6.0 & 0.62 & 0.78 & 152 & 4.30 & 10.9 \\
\rowcolor{teal!10}
\textbf{Claude 3.7-sonnet+DwT} & \textbf{0.87} & \textbf{0.95} & \textbf{39} & \textbf{5.12} & 3.6 & \textbf{0.75} & \textbf{0.91} & \textbf{57} & \textbf{4.93} & 11.2 & \textbf{0.70} & \textbf{0.86} & \textbf{85} & \textbf{4.72} & 24.9 \\
\bottomrule
\end{tabular}
\label{tab:model_comparison}
\end{table*}

\section{Experiment}

\subsection{Experimental Setup} 
To provide a comprehensive evaluation, we compared Draw-with-Thought (DwT) against eight state-of-the-art MLLMs— GPT-4o \cite{GPT-4o}, Claude3.5-sonnet \cite{Claude3.5Sonnet}, Claude3.7-sonnet \cite{anthropic2025claude37}, Gemini-2.0 \cite{Gemini2}, Grok-3 \cite{xai2025grok3}, Llama-3.2V-11B\cite{Llama3.2V} and Qwen2.5-VL-32B \cite{bai2025qwen2}.

Following \cite{rodriguez2023starvector,xing2024empowering}, we evaluate performance across two key dimensions using four metrics: \textbf{(1) Similarity}, measured by CLIP Score \cite{hessel2021clipscore, song2022clip}, DINO Score \cite{caron2021emerging, oquab2023dinov2}, and FID \cite{heusel2017gans}, to assess semantic and visual alignment with ground truth diagrams;
\textbf{(2) Aesthetics}, assessed using the Aesthetic Score \cite{aesthetic_christoph_2022}, which reflects perceived visual quality. Representative results are shown in Figure~\ref{fig:case}.

\paragraph{Overview. }We conducted comprehensive experiments evaluating our Draw with Thought (DwT) method against state-of-the-art MLMs on our Plot2XML benchmark across three difficulty levels. Our experimental results in Table~\ref{tab:model_comparison} reveal profound insights into the capabilities of MLLMs for scientific diagram reconstruction—a task that demands sophisticated visual-spatial reasoning, structural abstraction, and symbolic translation. Notably, our method consistently outperforms all baseline models across all metrics, with improvements of 10-20\% on semantic alignment scores and up to 40\% on visual quality metrics for the most challenging tasks. The performance patterns across models and difficulty levels illuminate the fundamental cognitive challenges in transforming rasterized diagrams into structured representations.

\paragraph{What drives our DwT superior performance? }Our method achieves remarkable improvements, particularly in Hard difficulty tasks, by fundamentally aligning with the cognitive processes underlying diagram comprehension\cite{abrahamsen2015diagrams,allen2011effects}. By guiding models through visual decomposition, structural inference, and symbolic code synthesis, our approach directly addresses the core challenges in scientific diagram parsing: maintaining coherent spatial-relational representations while translating between visual and symbolic modalities \cite{wu2023symbol}. The exceptional improvement in Hard tasks demonstrates that our method effectively scaffolds the complex reasoning required to handle diagrams with intricate spatial arrangements, nested hierarchies, and complex logical relationships—precisely the elements that make scientific diagrams challenging to reconstruct.

\paragraph{How does complexity impact performance? }The consistent performance degradation observed across all models as task difficulty increases directly reflects the escalating cognitive demands of complex diagram parsing. This degradation is particularly pronounced in base models without reasoning scaffolds, aligning with investigations of MLLM cognitive and memory mechanisms \cite{plass2010cognitive}, which indicate that complex visual-spatial tasks can overwhelm working memory when approached holistically \cite{srivastava2022beyond, lu2021fantastically}.  The disproportionate decline in FID scores compared to semantic alignment metrics suggests that models primarily struggle with maintaining spatial coherence and structural relationships rather than conceptual understanding.

\paragraph{What are current MLLMs' fundamental limitations? }Current MLLMs face a fundamental bottleneck in scientific diagram reconstruction: capturing multi-level abstractions often requires generating outputs exceeding the model's token limit, leading to truncation or content loss. While models like Claude 3.7-sonnet benefit from strong cross-modal consistency, their sharp performance drop on complex diagrams—reflected by a 26.8\% decline in CLIP score and a 105.5\% increase in FID from Easy to Hard tasks (Table~\ref{tab:model_comparison})—reveals the inadequacy of implicit spatial reasoning~\cite{ye2025longproc}. Conversely, the 47.2\% CLIP score drop observed in GPT-4o suggests that certain architectures fundamentally lack the capacity to preserve coherent spatial-relational representations across varying levels of diagrammatic complexity. Given that spatial arrangements encode underlying logical relationships, the inability to maintain such structure compromises semantic fidelity and highlights spatial-relational understanding as a critical yet underdeveloped competency in existing multimodal models.

\begin{figure*}[h!]
    \centering
    \includegraphics[width=0.9\textwidth]{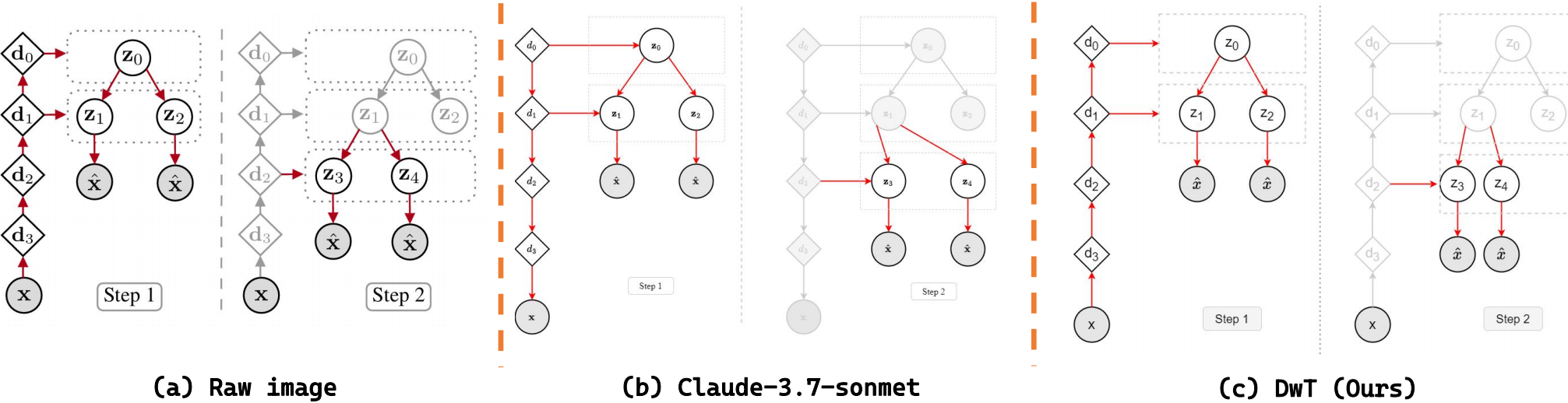}
    \caption{\textbf{Comparison of diagram reconstruction results} across different approaches.} 
    \label{human_eval}
\end{figure*}

\begin{table}[t]
\centering
\setlength{\tabcolsep}{4.5pt}
\caption{Ablation experiment. Results for Draw with Thought and ablation with different components using LLaMA 3.2-V-11B on Plot2XML.} 
\begin{tabular}{@{}l|cc|cc@{}}
\toprule
\textbf{Framework Variant} & DINO & CLIP & Aesth. & Valid. \\
\midrule
LLama 3.2-V + DwT& \textbf{0.815} & \textbf{0.657} & \textbf{3.807} & \textbf{89\%} \\
\midrule
\multicolumn{5}{@{}l}{\textit{Component Ablations:}} \\
w/o Perceptual Structuring & 0.742 & 0.610 & 3.726 & 85\% \\
w/o Layout Planning & 0.650 & 0.492 & 3.539 & 73\% \\
w/o Context Style Design & 0.795 & 0.582 & 3.541 & 87\% \\
\midrule
\multicolumn{5}{@{}l}{\textit{XML Generation Variants:}} \\
w/o Hierarchical XML & \multicolumn{2}{c|}{\textemdash} & \textemdash & 66\% \\
Flat XML + Refine & \multicolumn{2}{c|}{\textemdash} & \textemdash & 84\% \\
\bottomrule
\end{tabular}
\label{tab:ablation}
\end{table}

\subsection{Ablation Study}
Our Draw with Thought consists of three key components: \textit{Perceptual Structuring}, \textit{Semantic Layout Planning}, and \textit{Structured Code Generation}—which collectively support the extraction, organization, and formalization of visual content from scientific diagrams. The Context Style Design refers to the style extraction and definition step from our planning phase. To further enhance structural reliability, our framework incorporates a hierarchical XML generation module coupled with a self-refinement mechanism that iteratively corrects syntactic inconsistencies. Operates within an iterative feedback loop, acting as a targeted debugging way.  

As shown in Table~\ref{tab:ablation}, we conducted ablation studies on our Plot2XML benchmark using LLaMA 3.2-V-11B, evaluating CLIP and DINO for semantic alignment, aesthetic score for visual quality, and XML Validation for structural correctness. The results reveal a clear dependency chain, where errors in perceptual or planning stages cascade, impacting both semantic fidelity and structural integrity. This validates our design choice to separate high-level reasoning from low-level code generation, allowing for more robust and modular processing. These results demonstrate that each module contributes distinct functional value and that high-fidelity, machine-readable diagrams rely on the synergy of perceptual grounding, spatial reasoning, and structure-aware generation.

\subsection{Human Evaluation}
Following \cite{krosnick2017questionnaire,louviere2015best,flynn2014best}, we conducted a human evaluation to assess the alignment between automated metrics and human perception, combining absolute ratings and comparative judgments. For absolute ratings, domain experts used a 10-point scale~\cite{krosnick2017questionnaire} to evaluate each reconstructed diagram in terms of similarity to the original and aesthetic quality. For comparative evaluation, we employed Best-Worst Scaling (BWS)~\cite{louviere2015best,flynn2014best}, where annotators selected the best and worst outputs within a presented set. To manage annotation effort, comparisons were limited to the two strongest baseline models—Qwen2.5-VL and Claude 3.7-Sonnet—alongside our method (DWT), with human-annotated diagrams serving as reference. 

As shown in Table~\ref{tab:human_eval}, DWT achieves the highest scores in both similarity and aesthetics, closely approaching the human reference and outperforming the best LLMs. In BWS, DWT receives the highest score, indicating consistent preference, while Qwen’s negative score reflects frequent disfavor. These results confirm DWT’s superior semantic fidelity, visual quality, and alignment with human preferences.

\begin{table}[t!]
\centering
\renewcommand{\arraystretch}{1.2}
\caption{Human Evaluation Results. DWT surpasses all automated methods and closely matches human-created diagrams across all dimensions.}
\label{tab:human_eval}
\begin{tabular}{l|cccc}
\hline
\textbf{Metric} & \textbf{Qwen2.5V} & \textbf{Claude-3.7} & \textbf{DWT} & \textbf{Human} \\
\hline
Similarity & 5.6 & 6.1 & 7.3 & 9.4 \\
Aesthetics & 6.3 & 7.0 & 8.1 & 9.0 \\
BWS Score  & -0.87 & 0.13 & 0.74 & {N/A} \\
\hline
\end{tabular}
\end{table}

\paragraph{Correlation Analysis. }Spearman correlation between model metrics and human evaluations \cite{myers2014spearman} shows that CLIP scores exhibit the strongest correlation with human judgments of similarity (Spearman's $\rho = 0.825$), followed by DINO ($\rho = 0.763$) and FID ($\rho = -0.712$). This strong correlation with CLIP scores is particularly noteworthy, as it suggests that semantic alignment between the original and reconstructed diagrams—precisely what CLIP scores measure—is the most salient factor in human perception of reconstruction quality. The substantial negative correlation with FID scores indicates that while low-level visual fidelity matters, humans prioritize semantic and structural coherence over pixel-perfect reproduction when evaluating scientific diagrams.

\section{Conclusion}
In this paper, we present Draw with Thought, a training-free framework unleashing Multimodal Large Language Models (MLLMs) for reconstructing scientific diagrams from rasterized images into editable, executable mxGraph XML. Grounded in cognitive load \cite{plass2010cognitive} and structure mapping \cite{gentner1983structure} theory, our Draw with Thought involving Coarse-to-Fine Planning for
Structured Visual Understanding and Structure-Aware Code Generation via Progressive Realization—yields diagram representations that are both semantically faithful and structurally valid. To support systematic evaluation, we introduce Plot2XML, a benchmark of 247 real-world scientific diagrams with gold-standard XML annotations and multi-dimensional complexity analysis. We benchmarked our method across multiple state-of-the-art MLLMs, consistently demonstrating superior performance on high-complexity scientific diagrams through enhanced semantic alignment, layout fidelity, and XML validity.

In the future, we will further generalize our cognitively inspired pipeline to broader image-to-structure code generation tasks across scientific and industrial domains. The current framework is primarily evaluated on English-language diagrams from computer science publications, and its generalization to multilingual content or other academic domains remains to be explored.  Just as our training-free design avoids the prohibitive fine-tuning cost seen in \cite{xing2024empowering}, we also advocate for the development of efficient long-context adaptation strategies \cite{wu2024lotlip} for MLLMs—crucial for scaling symbolic reasoning under extended diagram code scenarios. Our Plot2XML benchmark offers a valuable testbed for evaluating such advances in context-efficient modeling tasks.
\begin{acks}
This work was supported in part by National Natural Science Foundation of China under contract No. 62303231 and in part by the Startup Foundation for Introducing Talent of NUIST under contract No. 2024r058. 
\end{acks}

\bibliographystyle{ACM-Reference-Format}
\bibliography{main}
\end{document}